\documentclass{article} 
\usepackage{iclr2017_workshop,times}
\usepackage{hyperref}
\usepackage{url}
\usepackage{amssymb}
\usepackage{amsmath}
\usepackage{algcompatible}
\usepackage{algorithm}
\usepackage{graphicx}

\usepackage{setspace}

\title{Multimodal Compact Bilinear Pooling for Multimodal Neural Machine Translation}

\author{Jean-Benoit Delbrouck \\
TCTS Lab\\
University of Mons, Belgium \\
\texttt{Jean-Benoit.DELBROUCK@umons.ac.be}
\And
Stephane Dupont\\
TCTS Lab\\
University of Mons, Belgium \\
\texttt{Stephane.DUPONT@umons.ac.be}
}

\begin{document}

\maketitle

\begin{abstract}
In state-of-the-art Neural Machine Translation, an attention mechanism is used during decoding to enhance the translation. At every step, the decoder uses this mechanism to focus on different parts of the source sentence to gather the most useful information before outputting its target word. Recently, the effectiveness of the attention mechanism has also been explored for multimodal tasks, where it becomes possible to focus both on sentence parts and image regions. Approaches to pool two modalities usually include element-wise product, sum or concatenation.
In this paper, we evaluate the more advanced Multimodal Compact Bilinear pooling method, which takes the outer product of two vectors to combine the attention features for the two modalities. This has been previously investigated for visual question answering. We try out this approach for multimodal image caption translation and show improvements compared to basic combination methods.
\end{abstract}

\section{Introduction}
In machine translation, neural networks have attracted a lot of research attention. Recently, the attention-based encoder-decoder framework~ \citep{SutskeverVL14,BahdanauCB14} has been largely adopted. In this approach, Recurrent Neural Networks (RNNs) map source sequences of words to target sequences. The attention mechanism is learned to focus on different parts of the input sentence while decoding. Attention mechanisms have been shown to work with other modalities too, like images, where their are able to learn to attend to salient parts of an image, for instance when generating text captions~\citep{XuBKCCSZB15}. For such  applications, Convolutional neural networks (CNNs) have shown to work best to represent images~\citep{He_2016_CVPR}.

Multimodal models of texts and images enable applications such as visual question answering or multimodal caption translation. Also, the grounding of multiple modalities against each other may enable the model to have a better understanding of each modality individually, such as in natural language understanding applications. 

The efficient integration of multimodal information still remains a challenging task though. Both \citet{huang2016attention} and \citet{caglayan2016does} made a first attempt in multimodal neural machine translation. Recently, \citet{calixto2017doubly} showed an improved architecture that significantly surpassed the monomodal baseline. Multimodal tasks require combining diverse modality vector representations with each other. Bilinear pooling models \citet{tenenbaum1997separating}, which computes the outer product of two vectors (such as the visual and textual representations), may be more expressive than basic combination methods such as element-wise sum or product. Because of its high and intractable dimensionality ($n^2$), \citet{gao2016compact} proposed a method that relies on Multimodal Compact Bilinear pooling (MCB) to efficiently compute a joint and expressive representation combining both modalities, in a visual question answering tasks. This approach has not been investigated previously for multimodal caption translation, which is what we focus on in this paper.

\begin{figure}[t]
  \begin{minipage}[t]{.475\textwidth}
    \vspace*{-\baselineskip}
   \label{TSalg}
\begin{algorithm}[H]
  \setstretch{1.1} 
  \caption{Multimodal CBP}
  \begin{algorithmic}[1]
    \State input: $v_1 \in \mathbb{R}^{n_1}, v_2 \in \mathbb{R}^{n_2}$
    \State output: $\Phi(v_1, v_2) \in \mathbb{R}^d$
    \State \textbf{for} $k \leftarrow 1 \hdots 2 $ \textbf{do}
    \State \qquad \textbf{for} $i \leftarrow 1 \hdots n_k$ \textbf{do}
    \State \qquad \qquad sample $h_k[i] $ from $\{1, \hdots,d\}$
    \State \qquad \qquad sample $s_k[i] $ from $\{-1, 1\}$
    \State \qquad $v'_k = \Psi(v_k,h_k,s_k,n_k)$
    \State \textbf{return} $\Phi = \text{FFT}^{-1}(\text{FFT}(v'_1) \odot \text{FFT}(v'_2))$
    \State \textbf{procedure} $\Psi(v,h,s,n)$
    \State \qquad \textbf{for} $i \hdots n$ \textbf{do}
    \State \qquad \qquad $y[h[i]] = y[h[i]] + s[i] \cdot v[i]$
    \State \qquad \textbf{return} $y$
  \end{algorithmic}
\end{algorithm}

  \end{minipage}%
  \hspace{4.75mm}
  \begin{minipage}[t]{.475\textwidth}
   \label{MCB}
    \centering
    \includegraphics[scale=0.55]{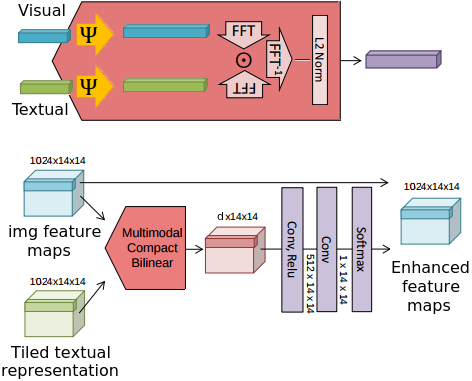}
    \vspace{-5mm}
  \end{minipage}
  \caption{Left: Tensor Sketch algorithm - Right: Compact Bilinear Pooling for two modality vectors (top) and \textit{"MM pre-attention"} model (bottom) ; Note that the textual representation vector is tiled (copied) to match the dimension of the image feature maps}
\end{figure}

\section{Model}
We detail our model build from the attention-based encoder-decoder neural network described by \citet{SutskeverVL14} and \citet{BahdanauCB14} implemented in \textit{TensorFlow}~\citep{abadi2016tensorflow}.
\textbf{Textual encoder} \quad Given an input sentence $X = (x_1, x_2, ..., x_T), x_i \in \mathbb{R}^E$ where $T$ is the sentence length and $E$ is the dimension of the word embedding space, a bi-directional LSTM encoder of layer size $L$ produces a set of textual annotation $A^{T} =  \{h^{t}_1, h^{t}_2, ...,h^{t}_T\}$ where $h_i$ is obtained by concatenating the forward and backward hidden states of the encoder: $h^{t}_i = [\overrightarrow{h_i};\overleftarrow{h_i}],h^{t}_i \in \mathbb{R}^{2L}$. \\
\textbf{Visual encoder} \quad An image associated to this sentence is fed to a deep residual network, computing convolutional feature maps of dimension $14\times14\times1024$. We obtain a set of visual annotations $A^{V} =  \{h^{v}_1, h^{v}_2, ...,h^{v}_{196}\}$ where $h^v_i \in {R}^{1024}$.\\ \\
\textbf{Decoder} \quad The decoder produces an output sentence $Y = (y_1, y_2, ..., y_{T'}), y_i \in \mathbb{R}^E$ and is initialized by $s_0 = tanh(W_{init} h^{t}_T + b_{init})$
where $h^{t}_T$ is the textual encoder's last state. The next decoder states are obtained as follows:
\begin{equation} s_t, o_t = LSTM(s_{t-1}, W_{in}[y_{t-1} ;c_{t-1}]), y_{t-1}  \in \mathbb{R}^E\end{equation}
During training, $y_{t-1}$ is the ground truth symbol in the sentence whilst $c_{t-1}$ is the previous attention vector computed by the attention model. The current attention vector $c_t$, concatenated with the LSTM output $o_t$, is used to compute a new vector $\tilde{o_t} = W_{proj} [o_t ; c_t] + b_{proj}$. The probability distribution over the target vocabulary is computed by the equation : \begin{equation} 
p(y_t|y_{t-1}, s_{t-1}, c_{t-1}, A^T, A^V) = softmax(W_{out} \tilde{o_t} + b_{out})
\end{equation}
\textbf{Attention} \quad At every time-step, the attention mechanism computes two modality specific context vectors  $\{c^t_t, c^v_t\}$ given the current decoder state $s_t$ and the two annotation sets $\{A^{T},A^{V}\}$. We use the same attention model for both modalities described by \citet{vinyals2015grammar}. We first compute modality specific attention weights $\alpha^{mod}_t = softmax(v^T tanh(W_1 A^{mod} + W_2 s_t + b))$. The context vector is then obtained with the following weighted sum : 
$c_t^{mod} = \sum\limits_{i=1}^{\lvert A^{mod} \rvert } \alpha_{ti}^{mod}h_i^{mod}$\\
Both $v^T$ and $W_1$ are considered modalities dependent and thus aren't shared by both modalities. The projection layer $W_2$ is applied to the decoder state $s_t$ and is thus shared~\citep{caglayan2016does}. Vectors $\{c^t_t, c^v_t\}$ are then combined to produce $c_t$ with an element-wise (e-w) sum / product or concatenation layer.  \\ \\
\textbf{Multimodal Compact Bilinear (MCB) pooling} \quad Bilinear models~\citep{tenenbaum1997separating} can be applied as vectors combination. We take the outer product of our two context vectors $c^{t}$ and $c^{v} \in \mathbb{R}^{2L}$ then learn a linear model $W$ i.e. $c_t = W [c^t_t \otimes  c^t_v]$, where $ \otimes$ denotes the outer product and $[$ $]$ denotes linearizing the matrix in a vector. Bilinear pooling allows all elements of both vectors to interact with each other in a multiplicative way but leads to a high dimensional representation and an infeasible number of parameters to learn in $W$. For two modality context vectors of size $2L = 1024$ and an attention size of $d = 512$ ($c_t \in \mathbb{R}^{512}$), $W$ would have $\approx $ 537 million parameters. We use the compact method proposed by \citet{gao2016compact}, based on the tensor sketch algorithm (see Algorithm \ref{TSalg}), to make bilinear models feasible. This model, referred as the \textit{"MM Attention"} in the results section, is illustrated in Figure \ref{MCB} (top right)

We try a second model inspired by the work of~\citep{fukui2016multimodal}. For each spatial grid location in the visual representation, we use MCB pooling to merge the slice of the visual feature with the language representation. As shown at the bottom right of Figure \ref{MCB}, after the pooling we use two convolutional layers to predict attention weights for each grid location. We then apply softmax to produce a new normalized soft attention map. This method can be seen as the removal of unnecessary information in the feature maps according to the source sentence. Note that we still use the \textit{"MM attention"} during decoding. We refer this model as the "\textit{MM pre-attention}".
\section{Settings}
We use the Adam optimizer~\citep{kingma2014adam} with a l.r. of $\alpha = 0.0007$ and L2 regularization of $\delta = 10^{-5}$. Layer size $L$ and word embeddings size $E$ is 512. Embeddings are trained along with the model. We use mini-batch size of 32 and Xavier weight initialization~\citep{glorot2010understanding}. For this experiments, we used the Multi30K dataset~\citep{elliott-EtAl:2016:VL16} which is an extended version of the Flickr30K Entities. For each image, one of the English descriptions was selected and manually translated into German by a professional translator (Task 1). As training and development data, 29,000 and 1,014 triples are used respectively. A test set of size 1000 is used for BLEU and METEOR evaluation. Vocabulary sizes are 11,180 (\textit{en}) and 19,154 (\textit{de}). We lowercase and tokenize all the text data with the Moses tokenizer. We extract feature maps from the images with a ResNet-50 at its $res4f\_relu$ layer. We use early-stopping if no improvement is observed after 10,000 steps.

\section{Results}
\vspace{-3mm}
\begin{table}[!htb]
  \caption{The BLEU and METEOR results on the test split containing 1000 triples. 		All scores are the average of two runs.}
  	\vspace{10px}
	\label{sample-table}
    \begin{minipage}{.400\textwidth}
      \begin{tabular}{lcc}
      \multicolumn{1}{c}{\bf Method}  &\multicolumn{2}{c}{\bf 				Validation Scores}
      \\ \hline \\
      &BLEU&METEOR \\
      Monomodal Text          &29.24&48.32 \\
      \textbf{MM attention} \\
      Concatenation            &26.12&44.14 \\
      Element-wise Sum         &27.48&45.79 \\
      Element-wise Product     &\textbf{28.62}&\textbf{47.99} \\
      MCB 1024                 &28.48&47.57 \\
      \hline \\
      \textbf{MM pre-attention*} \\
   Element-wise sum        &28.57&46.40\\
    Element-wise Product        &29.14&46.71 \\   
      MCB 4096        &\textbf{29.75}&\textbf{48.80} \\      
	\textit{ *with Prod as MM att.}
      \end{tabular}
    \end{minipage}%
      \hspace{20mm}
    \begin{minipage}{.400\textwidth}
      \centering
     \begin{tabular}{lll}
          \multicolumn{1}{c}{\bf Compact Bilinear $d$}  &\multicolumn{1}{c}{\bf BLEU}
          \\ \hline \\
          \textbf{Multimodal attention} \\
          512         &27.78\\
          1024         &\textbf{28.48}\\
          2048          &28.12\\
          \hline \\
          \textbf{Multimodal pre-attention} \\
          1024            &28.71\\
          2048            &29.19\\
          4096            &\textbf{29.75}\\
          8192            &29.39\\
          16000           &27.98\\

          \end{tabular}
    \end{minipage} 
\end{table}

To our knowledge, there is currently no multimodal translation architecture that convincingly surpass a monomodal NMT baseline. Our work nevertheless shows a small but encouraging improvement. In the \textit{"MM attention"} model, where both attention context vectors are merged, we notice no improvement using MCB over an element-wise product. We suppose the reason is that the merged attention vector $c_t$ has to be concatenated with the cell output and then gets linearly transformed by the \textit{proj} layer to a vector of size \textit{512}. This heavy dimensionality reduction undergone by the vector may have lead to a consequent loss of information, thus the poor results. This motivated us to implement the second attention mechanism, \textit{"MM pre-attention"}. Here, the attention model can enjoy the full use of the combined vectors dimension, varying from 1024 to 16000. We show here an improvement of +0.62 BLEU over e-w multiplication and +1.18 BLEU over e-w sum. We believe a step further could be to investigate different experimental settings or layer architectures as we felt MCB could perform much better as seen in similar previous work~\citep{fukui2016multimodal}.
%
%
%

\section{Acknowledgements}
This work was partly supported by the Chist-Era project IGLU with contribution from the Belgian Fonds de la Recherche Scientique (FNRS), contract no. R.50.11.15.F, and by the FSO project VCYCLE with contribution from the Belgian Waloon Region, contract no. 1510501.

\bibliography{iclr2017_workshop}
\bibliographystyle{iclr2017_workshop}

\end{document}